\documentclass{article}
\usepackage{spconf,amsmath,amssymb,graphicx,multirow}

\title{Learning Representations of Emotional Speech with Deep Convolutional Generative Adversarial Networks}
\twoauthors
  {Jonathan Chang}
	{Harvey Mudd College\\
	Computer Science Department\\
	1250 North Dartmouth Ave\\
    Claremont, CA 91711}
  {Stefan Scherer\thanks{This material is based upon work supported by the U.S. Army Research Laboratory under contract number W911NF-14-D-0005. Any opinions, findings, and conclusions or recommendations expressed in this material are those of the author(s) and do not necessarily reflect the views of the Government, and no official endorsement should be inferred.}}
    {University of Southern California\\
	Institute for Creative Technologies\\
	12015 Waterfront Dr\\
    Playa Vista, CA 90094}

\begin{document}
\ninept
\maketitle
\begin{abstract}
Automatically assessing emotional valence in human speech has historically been a difficult task for machine learning algorithms. The subtle changes in the voice of the speaker that are indicative of positive or negative emotional states are often "overshadowed" by voice characteristics relating to emotional intensity or emotional activation. In this work we explore a representation learning approach that automatically derives discriminative representations of emotional speech. In particular, we investigate two machine learning strategies to improve classifier performance: (1) utilization of unlabeled data using a deep convolutional generative adversarial network (DCGAN), and (2) multitask learning. Within our extensive experiments we leverage a multitask annotated emotional corpus as well as a large unlabeled meeting corpus (around 100 hours). Our speaker-independent classification experiments show that in particular the use of unlabeled data in our investigations improves performance of the classifiers and both fully supervised baseline approaches are outperformed considerably. We improve the classification of emotional valence on a discrete 5-point scale to 43.88\% and on a 3-point scale to 49.80\%, which is competitive to state-of-the-art performance.
\end{abstract}
\section{Introduction}
\label{sec:intro}

Machine Learning, in general, and affective computing, in particular, rely on good data representations or features that have a good discriminatory faculty in classification and regression experiments, such as emotion recognition from speech. To derive efficient representations of data, researchers have adopted two main strategies: (1) carefully crafted and tailored feature extractors designed for a particular task \cite{dollar2007feature} and (2) algorithms that learn representations automatically from the data itself \cite{bengio2013representation}. The latter approach is called Representation Learning (RL), and has received growing attention in the past few years and is highly reliant on large quantities of data. Most approaches for emotion recognition from speech still rely on the extraction of standard acoustic features such as pitch, shimmer, jitter and MFCCs (Mel-Frequency Cepstral Coefficients), with a few notable exceptions \cite{trigeorgis2016adieu,DBLP:journals/corr/GhoshLMS15,GhoshetAl_2016,GhoshetAl_2016b}. In this work we leverage RL strategies and automatically learn representations of emotional speech from the spectrogram directly using a deep convolutional neural network (CNN) architecture. 

To learn strong representations of speech we seek to leverage as much data as possible. However, emotion annotations are difficult to obtain and scarce \cite{mckeown2012semaine}. We leverage the USC-IEMOCAP dataset, which comprises of around 12 hours of highly emotional and partly acted data from 10 speakers \cite{busso2008iemocap}. However, we aim to improve the learned representations of emotional speech with unlabeled speech data from an unrelated meeting corpus, which consists of about 100 hours of data \cite{carletta2005ami}. While the meeting corpus is qualitatively quite different from the highly emotional USC-IEMOCAP data\footnote{The AMI meeting corpus is not highly emotional and strong emotions such as anger do not appear.}, we believe that the learned representations will improve through the use of these additional data. This combination of two separate data sources leads to a semi-supervised machine learning task and we extend the CNN architecture to a deep convolutional generative neural network (DCGAN) that can be trained in an unsupervised fashion \cite{radford2015unsupervised}. 

Within this work, we particularly target emotional valence as the primary task, as it has been shown to be the most challenging emotional dimension for acoustic analyses in a number of studies \cite{DangetAl_2016,busso2012unveiling}. Apart from solely targeting valence classification, we further investigate the principle of multitask learning. In multitask learning, a set of related tasks are learned (e.g., emotional activation), along with a primary task (e.g., emotional valence); both tasks share parts of the network topology and are hence jointly trained, as depicted in Figure \ref{fig:nn}. It is expected that data for the secondary task models information, which would also be discriminative in learning the primary task. In fact, this approach has been shown to improve generalizability across corpora \cite{ghosh2015multi}.

The remainder of this paper is organized as follows: First we introduce the DCGAN model and discuss prior work, in Section \ref{sec:prior}. Then we describe our specific multitask DCGAN model in Section \ref{sec:model}, introduce the datasets in Section \ref{sec:corpus}, and describe our experimental design in Section \ref{sec:experiment}. Finally, we report our results in Section \ref{sec:results} and discuss our findings in Section \ref{sec:conclusions}.

\section{Related Work}
\label{sec:prior}

The proposed model builds upon previous results in the field of emotion recognition, and leverages prior work in representation learning.

Multitask learning has been effective in some prior experiments on emotion detection. In particular, Xia and Liu proposed a multitask model for emotion recognition which, like the investigated model, has activation and valence as targets \cite{xia_liu}. Their work uses a Deep Belief Network (DBN) architecture to classify the emotion of audio input, with valence and activation as secondary tasks. Their experiments indicate that the use of multitask learning produces improved unweighted accuracy on the emotion classification task. Like Xia and Liu, the proposed model uses multitask learning with valence and activation as targets. Unlike them, however, we are primarily interested not in emotion classification, but in valence classification as a primary task. Thus, our multitask model has valence as a primary target and activation as a secondary target. Also, while our experiments use the IEMOCAP database like Xia and Liu do, our method of speaker split differs from theirs. Xia and Liu use a leave-one-speaker-out cross validation scheme with separate train and test sets that have no speaker overlap. This method lacks a distinct validation set; instead, they validate and test on the same set. Our experimental setup, on the other hand, splits the data into distinct train, validation, and test sets, still with no speaker overlap. This is described in greater detail in Section \ref{sec:experiment}.

The unsupervised learning part of the investigated model builds upon an architecture known as the deep convolutional generative adversarial network, or DCGAN. DCGAN consists of two components, known as the \textit{generator} and \textit{discriminator}, which are trained against each other in a minimax setup. The generator learns to map samples from a random distribution to output matrices of some pre-specified form. The discriminator takes an input which is either a generator output or a ``real'' sample from a dataset. The discriminator learns to classify the input as either generated or real \cite{radford2015unsupervised}.

For training, the discriminator uses a cross entropy loss function based on how many inputs were correctly classified as real and how many were correctly classified as generated. The cross entropy loss between true labels $y$ and predictions $\hat{y}$ is defined as:
\begin{equation}
\label{eq1}
\mathcal{L}(\mathbf{w}) = -\frac{1}{N}\sum_{n=1}^N \left[ y_n\log\hat{y}_n + (1-y_n)\log(1-\hat{y}_n) \right]
\end{equation}
Where $\mathbf{w}$ is the learned vector of weights, and $N$ is the number of samples. For purposes of this computation, labels are represented as numerical values of 1 for real and 0 for generated. Then, letting $\hat{y}_r$ represent the discriminator's predictions for all real inputs, the cross entropy for correct predictions of ``real'' simplifies to:
\[
\mathcal{L}_r(\mathbf{w}) = -\frac{1}{N}\sum_{n=1}^N \log\hat{y}_{r,n}
\]
Because in this case the correct predictions are all ones. Similarly, letting $\hat{y}_g$ represent the discriminator's predictions for all generated inputs, the cross entropy for correct predictions of ``generated'' simplifies to:
\[
\mathcal{L}_f(\mathbf{w}) = -\frac{1}{N}\sum_{n=1}^N \log(1-\hat{y}_{g,n})
\]
Because here the correct predictions are all zeroes. The total loss for the discriminator is given by the sum of the previous two terms $\mathcal{L}_d = \mathcal{L}_r + \mathcal{L}_f$.

The generator also uses a cross entropy loss, but its loss is defined in terms of how many generated outputs got \emph{incorrectly} classified as real:
\[
\mathcal{L}_g(\mathbf{w}) = -\frac{1}{N}\sum_{n=1}^N \log(\hat{y}_{g,n})
\]
Thus, the generator's loss gets lower the better it is able to produce outputs that the discriminator thinks are real. This leads the generator to eventually produce outputs that look like real samples of speech given sufficient training iterations.

\section{Multitask Deep Convolutional Generative Adversarial Network}
\label{sec:model}

\begin{figure}[htb]
\centering
\includegraphics[width=8.0cm]{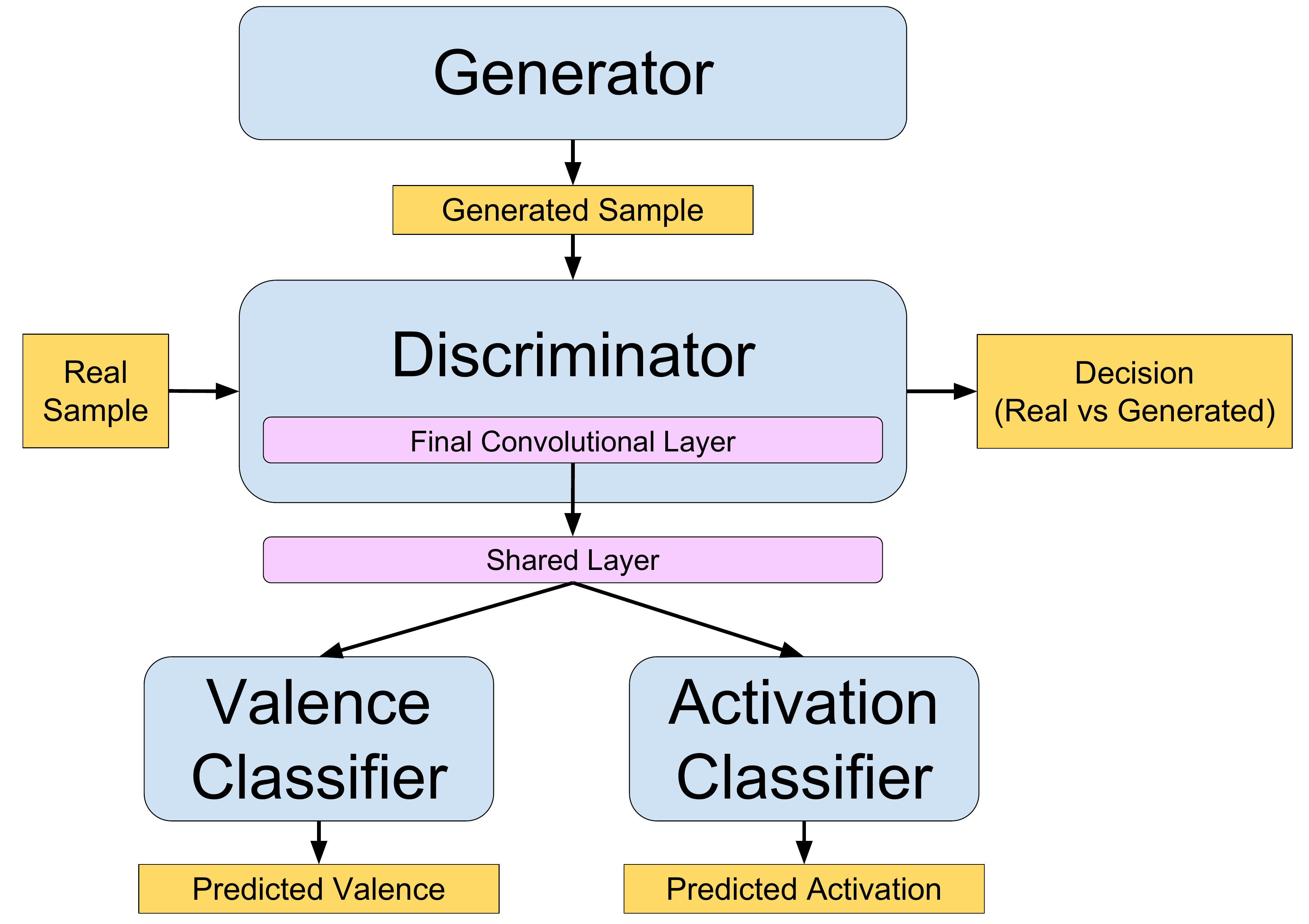}
\caption{Visual representation of the deep convolutional generative adversarial network with multitask valence and activation classifier.}
\label{fig:nn}
\end{figure}

The investigated multitask model is based upon the DCGAN architecture described in Section \ref{sec:prior} and is implemented in TensorFlow\footnote{We will make the code available on GitHub upon publication.}. For emotion classification a fully connected layer is attached to the final convolutional layer of the DCGAN's discriminator. The output of this layer is then fed to two separate fully connected layers, one of which outputs a valence label and the other of which outputs an activation label. This setup is shown visually in Figure \ref{fig:nn}. Through this setup, the model is able to take advantage of unlabeled data during training by feeding it through the DCGAN layers in the model, and is also able to take advantage of multitask learning and train the valence and activation outputs simultaneously.

In particular, the model is trained by iteratively running the generator, discriminator, valence classifier, and activation classifier, and back-propagating the error for each component through the network. The loss functions for the generator and discriminator are unaltered, and remain as shown in Section \ref{sec:prior}. Both the valence classifier and activation classifier use cross entropy loss as in Equation \ref{eq1}.

Since the valence and activation classifiers share layers with the discriminator the model learns features and convolutional filters that are effective for the tasks of valence classification, activation classification, and discriminating between real and generated samples. 

\section{Data Corpus}
\label{sec:corpus}

Due to the semi-supervised nature of the proposed Multitask DCGAN model, we utilize both labeled and unlabeled data. For the unlabeled data, we use audio from the AMI \cite{carletta2005ami} and IEMOCAP \cite{busso2008iemocap} datasets. For the labeled data, we use audio from the IEMOCAP dataset, which comes with labels for activation and valence, both measured on a 5-point Likert scale from three distinct annotators. Although IEMOCAP provides per-word activation and valence labels, in practice these labels do not generally change over time in a given audio file, and so for simplicity we label each audio clip with the \emph{average} valence and activation. Since valence and activation are both measured on a 5-point scale, the labels are encoded in 5-element one-hot vectors. For instance, a valence of 5 is represented with the vector $[0,0,0,0,1]$. The one-hot encoding can be thought of as a probability distribution representing the likelihood of the correct label being some particular value. Thus, in cases where the annotators disagree on the valence or activation label, this can be represented by assigning probabilities to multiple positions in the label vector. For instance, a label of 4.5 conceptually means that the ``correct'' valence is either 4 or 5 with equal probability, so the corresponding vector would be $[0,0,0,0.5,0.5]$. These ``fuzzy labels'' have been shown to improve classification performance in a number of applications \cite{IFIFOSVMRVQC,FIFOOAASVM}. It should be noted here that we had generally greater success with this fuzzy label method than training the neural network model on the valence label directly, i.e. classification task vs. regression.

\textbf{Pre-processing.} Audio data is fed to the network models in the form of spectrograms. The spectrograms are computed using a short time Fourier transform with window size of 1024 samples, which at the 16 kHz sampling rate is equivalent to 64 ms. Each spectrogram is 128 pixels high, representing the frequency range 0-11 kHz. Due to the varying lengths of the IEMOCAP audio files, the spectrograms vary in width, which poses a problem for the batching process of the neural network training. To compensate for this, the model randomly crops a region of each input spectrogram. The crop width is determined in advance. To ensure that the selected crop region contains at least some data (i.e. is not entirely silence), cropping occurs using the following procedure: a random word in the transcript of the audio file is selected, and the corresponding time range is looked up. A random point within this time range is selected, which is then treated as the center line of the crop. The crop is then made using the region defined by the center line and crop width.

Early on, we found that there is a noticeable imbalance in the valence labels for the IEMOCAP data, in that the labels skew heavily towards the neutral (2-3) range. In order to prevent the model from overfitting to this distribution during training, we \textit{normalize} the training data by oversampling underrepresented valence data, such that the overall distribution of valence labels is more even.

\section{Experimental Design}
\label{sec:experiment}

\textbf{Investigated Models.} We investigate the impact of both unlabeled data for improved emotional speech representations and multitask learning on emotional valence classification performance. To this end, we compared four different neural network models:
\begin{enumerate}
	\item \textbf{BasicCNN:} A fully supervised, single-task valence classifier.
    \item \textbf{MultitaskCNN:} A fully supervised, multitask valence classifier with activation as secondary task.
    \item \textbf{BasicDCGAN:} A semi-supervised, DCGAN with single-task valence classifier.
    \item \textbf{MultitaskDCGAN:} A semi-supervised, DCGAN with multitask valence classifier and activation as secondary task.
\end{enumerate}

The BasicCNN represents a ``bare minimum'' valence classifier and thus sets a lower bound for expected performance. Comparison with MultitaskCNN indicates the effect of the inclusion of a secondary task, i.e. emotional activation recognition. Comparison with BasicDCGAN indicates the effect of the incorporation of unlabeled data during training.

For fairness, the architectures of all three baselines are based upon the full MultitaskDCGAN model. BasicDCGAN for example is simply the MultitaskDCGAN model with the activation layer removed, while the two fully supervised baselines were built by taking the convolutional layers from the discriminator component of the MultitaskDCGAN, and adding fully connected layers for valence and activation output. Specifically, the discriminator contains four convolutional layers; there is no explicit pooling but the kernel stride size is 2 so image size gets halved at each step. Thus, by design, all four models have this same convolutional structure. This is to ensure that potential performance gains do not stem from a larger complexity or higher number of trainable weights within the DCGAN models, but rather stem from improved representations of speech.

\textbf{Experimental Procedure.} The parameters for each model, including batch size, filter size (for convolution), and learning rate, were determined by randomly sampling different parameter combinations, training the model with those parameters, and computing accuracy on a held-out validation set. For each model, we kept the parameters that yield the best accuracy on the held-out set. This procedure ensures that each model is fairly represented during evaluation. Our hyper-parameters included crop width of the input signal $\in \{64, 128\}$, convolutional layer filter sizes $\in [2, w/8]$ (where $w$ is the selected crop width and gets divided by 8 to account for each halving of image size from the 3 convolutional layers leading up to the last one), number of convolutional filters $\in [32, 90]$ (step size 4), batch size $\in \{64, 128, 256\}$, and learning rates $\in \{1e-3, 1e-4, 1e-5\}$. Identified parameters per model are shown in Table~\ref{tab:parameters}.

\begin{table}[!t]
\centering
\begin{tabular}{lcccc}
\hline
~ & Basic & Multitask & Basic & Multitask \\
Parameter & CNN & CNN & DCGAN & DCGAN \\
\hline
\hline
Crop Width & 128 & 64 & 128 & 64 \\
Learning Rate & $1e-4$ & $1e-3$ & $1e-4$ & $1e-4$ \\
Batch Size & 64 & 128 & 64 & 128 \\
Filter Size & 15 & 8 & 9 & 6 \\
No. of Filters & 84 & 32 & 72 & 88 \\
\hline
\end{tabular}
\caption{Final parameters used for each model as found by random parameter search}
\label{tab:parameters}
\end{table}

For evaluation, we utilized a 5-fold leave-one-session-out validation. Each fold leaves one of the five sessions in the labeled IEMOCAP data out of the training set entirely. From this left-out conversation, one speaker's audio is used as a validation set, while the other speaker's audio is used as a test set. 

For each fold, the evaluation procedure is as follows: the model being evaluated is trained on the training set, and after each full pass through the training set, accuracy is computed on the validation set. This process continues until the accuracy on the validation set is found to no longer increase; in other words, we locate a local maximum in validation accuracy. To increase the certainty that this local maximum is truly representative of the model's best performance, we continue to run more iterations after a local maximum is found, and look for 5 consecutive iterations with lower accuracy values. If, in the course of these 5 iterations, a higher accuracy value is found, that is treated as the new local maximum and the search restarts from there. Once a best accuracy value is found in this manner, we restore the model's weights to those of the iteration corresponding to the best accuracy, and evaluate the accuracy on the test set.

We evaluated each model on all 5 folds using the methodology described above, recording test accuracies for each fold.

\textbf{Evaluation Strategy.} We collected several statistics about our models' performances. We were primarily interested in the unweighted per-class accuracy. In addition, we converted the network's output from probability distributions back into numerical labels by taking the expected value; that is:
$
v = \sum_{i=1}^5 ip_i,
$
where $p$ is the model's prediction, in its original form as a 5 element vector probability distribution. We then used this to compute the Pearson correlation ($\rho$ measure) between predicted and actual labels.

Some pre-processing was needed to obtain accurate measures. In particular, in cases where human annotators were perfectly split on what the correct label for a particular sample should be, both possibilities should be accepted as correct predictions. For instance, if the correct label is 4.5 (vector form $[0,0,0,0.5,0.5]$), a correct prediction could be either 4 \emph{or} 5 (i.e. the maximum index in the output vector should either be 4 or 5). 

The above measures are for a 5-point labeling scale, which is how the IEMOCAP data is originally labeled. However, prior experiments have evaluated performance on valence classification on a 3-point scale \cite{metallinou2012context}. The authors provide an example of this, with valence levels 1 and 2 being pooled into a single ``negative'' category, valence level 3 remaining untouched as the ``neutral'' category, and valence levels 4 and 5 being pooled into a single ``positive'' category. Thus, to allow for comparison with our models, we also report results on such a 3-point scale. We construct these results by taking the results for 5 class comparison and pooling them as just described.

\section{Results}
\label{sec:results}

Table \ref{tab:accs} shows the unweighted per-class accuracies and Pearson correlation coeffecients ($\rho$ values) between actual and predicted labels for each model. All values shown are average values across the test sets for all 5 folds.

\begin{table}[!t]
\centering
\begin{tabular}{lccc}
\hline
    ~ & Accuracy & Accuracy & Pearson Correlation \\
    Model & (5 class) & (3 class) & ($\rho$ value) \\
    \hline \hline
    BasicCNN & $38.52\%$ & $46.59\%$ & $0.1639$ \\
    MultitaskCNN & $36.78\%$ & $40.57\%$ & $0.0737$ \\
    \textbf{BasicDCGAN} & $\mathbf{43.88\%}$ & $\mathbf{49.80\%}$ & $\mathbf{0.2618}$ \\
    MultitaskDCGAN & $43.69\%$ & $48.88\%$ & $0.2434$ \\ \hline
\end{tabular}
\caption{Evaluation metrics for all four models, averaged across 5 test folds. Speaker-independent unweighted accuracies in \% for both 5-class and 3-class valence performance as well as Pearson correlation $\rho$ are reported.}
\label{tab:accs}
\end{table}

Results indicate that the use of unsupervised learning yields a clear improvement in performance. Both BasicDCGAN and MultitaskDCGAN have considerably better accuracies and linear correlations compared to the fully supervised CNN models. This is a strong indication that the use of large quantities of task-unrelated speech data improved the filter learning in the CNN layers of the DCGAN discriminator. 

Multitask learning, on the other hand, does not appear to have any positive impact on performance. Comparing the two CNN models, the addition of multitask learning actually appears to impair performance, with MultitaskCNN doing worse than BasicCNN in all three metrics. The difference is smaller when comparing BasicDCGAN and MultitaskDCGAN, and may not be enough to decidedly conclude that the use of multitask learning has a net negative impact there, but certainly there is no indication of a net positive impact. The observed performance of both the BasicDCGAN and MultitaskDCGAN using 3-classes is comparable to the state-of-the-art, with 49.80\% compared to 49.99\% reported in \cite{metallinou2012context}. It needs to be noted that in \cite{metallinou2012context} data from the test speaker's session partner was utilized in the training of the model. Our models in contrast are trained on only four of the five sessions as discussed in \ref{sec:experiment}. Further, the here presented models are trained on the raw spectrograms of the audio and no feature extraction was employed whatsoever. This representation learning approach is employed in order to allow the DCGAN component of the model to train on vast amounts of unsupervised speech data.

We further report the confusion matrix of the best performing model BasicDCGAN in Table \ref{tab:confmat}. It is noted that the ``negative'' class (i.e., the second row) is classified the best. However, it appears that this class is picked more frequently by the model resulting in high recall = 0.7701 and low precision = 0.3502. The class with the highest F1 score is ``very positive'' (i.e., the last row) with $F1 = 0.5284$. The confusion of ``very negative'' valence with ``very positive'' valence in the top right corner is interesting and has been previously observed \cite{GhoshetAl_2016}.

\section{Conclusions}
\label{sec:conclusions}

We investigated the use of unsupervised and multitask learning to improve the performance of an emotional valence classifier. Overall, we found that unsupervised learning yields considerable improvements in classification accuracy for the emotional valence recognition task. The best performing model achieves 43.88\% in the 5-class case and 49.80\% in the 3-class case with a significant Pearson correlation between continuous target label and prediction of $\rho = 0.2618$ ($p < .001$). There is no indication that multitask learning provides any advantage.

\begin{table}[!t]
\centering
\begin{tabular}{lccccc}
\hline
    & Very Neg. & Neg. & Neu. & Pos. & Very Pos. \\
    \hline \hline
    Very Neg. &\textbf{0.3316}  &  0.1633 &   0.1837  &  0.2092  &  0.1122\\
    Neg. &0.0293  &  \textbf{0.7701} &   0.0796  &  0.0910  &  0.0299\\
     Neu. &0.0106  &  0.5090 &   \textbf{0.3507}  &  0.0903  &  0.0393\\
   Pos. &0.0285  &  0.4881 &   0.1395  &  \textbf{0.3027} &  0.0412\\
    Very Pos. &0.0366  &  0.2683 &   0.0732  &  0.1829  &  \textbf{0.4390}\\ \hline
\end{tabular}
\caption{Confusion matrix for 5-class valence classification with the BasicDCGAN model. Predictions are reported in columns and actual targets in rows. Valence classes are sorted from very negative to very positive. These classes correspond to the numeric labels 1 through 5.}
\label{tab:confmat}
\end{table}

The results for multitask learning are somewhat surprising. It may be that the valence and activation classification tasks are not sufficiently related for multitask learning to yield improvements in accuracy. Alternatively, a different neural network architecture may be needed for multitask learning to work. Further, the alternating update strategy employed in the present work might not have been the optimal strategy for training. The iterative swapping of target tasks valence/activation might have created instabilities in the weight updates of the backpropagation algorithm. There may yet be other explanations; further investigation may be warranted.

Lastly, it is important to note that this model's performance is only approaching state-of-the-art, which employs potentially better suited sequential classifiers such as Long Short-term Memory (LSTM) networks \cite{hochreiter1997long}. However, basic LSTM are not suited to learn from entirely unsupervised data, which we leveraged for the proposed DCGAN models. For future work, we hope to adapt the technique of using unlabeled data to sequential models, including LSTM. We expect that combining our work here with the advantages of sequential models may result in further performance gains which may be more competitive with today's leading models and potentially outperform them. For the purposes of this investigation, the key takeaway is that the use of unsupervised learning yields clear performance gains on the emotional valence classification task, and that this represents a technique that may be adapted to other models to achieve even higher classification accuracies.


\vfill\pagebreak

\bibliographystyle{IEEEbib}
\bibliography{refs}

\begin{thebibliography}{10}

\bibitem{dollar2007feature}
Piotr Doll{\'a}r, Zhuowen Tu, Hai Tao, and Serge Belongie,
\newblock ``Feature mining for image classification,''
\newblock in {\em 2007 IEEE Conference on Computer Vision and Pattern
  Recognition}. IEEE, 2007, pp. 1--8.

\bibitem{bengio2013representation}
Yoshua Bengio, Aaron Courville, and Pascal Vincent,
\newblock ``Representation learning: A review and new perspectives,''
\newblock {\em IEEE transactions on pattern analysis and machine intelligence},
  vol. 35, no. 8, pp. 1798--1828, 2013.

\bibitem{trigeorgis2016adieu}
George Trigeorgis, Fabien Ringeval, Raymond Brueckner, Erik Marchi, Mihalis~A
  Nicolaou, Stefanos Zafeiriou, et~al.,
\newblock ``Adieu features? end-to-end speech emotion recognition using a deep
  convolutional recurrent network,''
\newblock in {\em 2016 IEEE International Conference on Acoustics, Speech and
  Signal Processing (ICASSP)}. IEEE, 2016, pp. 5200--5204.

\bibitem{DBLP:journals/corr/GhoshLMS15}
Sayan Ghosh, Eugene Laksana, Louis{-}Philippe Morency, and Stefan Scherer,
\newblock ``Learning representations of affect from speech,''
\newblock {\em CoRR}, vol. abs/1511.04747, 2015.

\bibitem{GhoshetAl_2016}
S.~Ghosh, E.~Laksana, L.-P. Morency, and S.~Scherer,
\newblock ``Representation learning for speech emotion recognition,''
\newblock in {\em Proceedings of Interspeech 2016}, 2016.

\bibitem{GhoshetAl_2016b}
S.~Ghosh, L.-P. Morency, E.~Laksana, and S.~Scherer,
\newblock ``An unsupervised approach to glottal inverse filtering,''
\newblock in {\em Proceedings of EUSIPCO 2016}, 2016.

\bibitem{mckeown2012semaine}
Gary McKeown, Michel Valstar, Roddy Cowie, Maja Pantic, and Marc Schroder,
\newblock ``The semaine database: Annotated multimodal records of emotionally
  colored conversations between a person and a limited agent,''
\newblock {\em IEEE Transactions on Affective Computing}, vol. 3, no. 1, pp.
  5--17, 2012.

\bibitem{busso2008iemocap}
Carlos Busso, Murtaza Bulut, Chi-Chun Lee, Abe Kazemzadeh, Emily Mower, Samuel
  Kim, Jeannette~N Chang, Sungbok Lee, and Shrikanth~S Narayanan,
\newblock ``Iemocap: Interactive emotional dyadic motion capture database,''
\newblock {\em Language resources and evaluation}, vol. 42, no. 4, pp.
  335--359, 2008.

\bibitem{carletta2005ami}
Jean Carletta, Simone Ashby, Sebastien Bourban, Mike Flynn, Mael Guillemot,
  Thomas Hain, Jaroslav Kadlec, Vasilis Karaiskos, Wessel Kraaij, Melissa
  Kronenthal, et~al.,
\newblock ``The ami meeting corpus: A pre-announcement,''
\newblock in {\em International Workshop on Machine Learning for Multimodal
  Interaction}. Springer, 2005, pp. 28--39.

\bibitem{radford2015unsupervised}
Alec Radford, Luke Metz, and Soumith Chintala,
\newblock ``Unsupervised representation learning with deep convolutional
  generative adversarial networks,''
\newblock {\em arXiv preprint arXiv:1511.06434}, 2015.

\bibitem{DangetAl_2016}
T.~Dang, V.~Sethu, and E.~Ambikairajah,
\newblock ``Factor analysis based speaker normalisation for continuous emotion
  prediction,''
\newblock in {\em Proceedings of Interspeech 2016}, 2016.

\bibitem{busso2012unveiling}
Carlos Busso and Tauhidur Rahman,
\newblock ``Unveiling the acoustic properties that describe the valence
  dimension.,''
\newblock in {\em INTERSPEECH}, 2012, pp. 1179--1182.

\bibitem{ghosh2015multi}
Sayan Ghosh, Eugene Laksana, Stefan Scherer, and Louis-Philippe Morency,
\newblock ``A multi-label convolutional neural network approach to cross-domain
  action unit detection,''
\newblock in {\em Affective Computing and Intelligent Interaction (ACII), 2015
  International Conference on}. IEEE, 2015, pp. 609--615.

\bibitem{xia_liu}
Rui Xia and Yang Liu,
\newblock ``A multi-task learning framework for emotion recognition using 2d
  continuous space,''
\newblock {\em IEEE Transactions on Affective Computing}, 2015.

\bibitem{IFIFOSVMRVQC}
S.~Scherer, J.~Kane, C.~Gobl, and F.~Schwenker,
\newblock ``Investigating fuzzy-input fuzzy-output support vector machines for
  robust voice quality classification,''
\newblock {\em Computer Speech and Language}, vol. 27, no. 1, pp. 263--287,
  2013.

\bibitem{FIFOOAASVM}
C.~Thiel, S.~Scherer, and F.~Schwenker,
\newblock ``Fuzzy-input fuzzy-output one-against-all support vector machines,''
\newblock in {\em 11th International Conference on Knowledge-Based and
  Intelligent Information and Engineering Systems (KES'07)}. 2007, vol.~3 of
  {\em Lecture Notes in Artificial Intelligence}, pp. 156--165, Springer.

\bibitem{metallinou2012context}
Angeliki Metallinou, Martin Wollmer, Athanasios Katsamanis, Florian Eyben,
  Bjorn Schuller, and Shrikanth Narayanan,
\newblock ``Context-sensitive learning for enhanced audiovisual emotion
  classification,''
\newblock {\em IEEE Transactions on Affective Computing}, vol. 3, no. 2, pp.
  184--198, 2012.

\bibitem{hochreiter1997long}
Sepp Hochreiter and J{\"u}rgen Schmidhuber,
\newblock ``Long short-term memory,''
\newblock {\em Neural computation}, vol. 9, no. 8, pp. 1735--1780, 1997.

\end{thebibliography}

\end{document}